\documentclass{article}

\usepackage{microtype}
\usepackage{graphicx}
\usepackage{subfigure}
\usepackage{booktabs} 

\usepackage{hyperref}

\usepackage[accepted]{icml2023}

\usepackage{amsmath}
\usepackage{amssymb}
\usepackage{mathtools}
\usepackage{amsthm}

\usepackage[capitalize,noabbrev]{cleveref}

\theoremstyle{plain}

\theoremstyle{definition}

\theoremstyle{remark}

\usepackage[textsize=tiny]{todonotes}

\icmltitlerunning{}

\begin{document}

\twocolumn[
\icmltitle{A Denoising Diffusion Model for Fluid Field Prediction}



\icmlsetsymbol{equal}{*}

\begin{icmlauthorlist}
\icmlauthor{Gefan Yang}{equal,yyy}
\icmlauthor{Stefan Sommer}{equal,xxx}
\end{icmlauthorlist}

\icmlaffiliation{yyy}{Niels Bohr Institute, University of Copenhagen, Copenhagen, Denmark}
\icmlaffiliation{xxx}{DIKU, University of Copenhagen, Copenhagen, Denmark}

\icmlcorrespondingauthor{Gefan Yang}{vbd402@alumni.ku.dk}

\icmlkeywords{Machine Learning, ICML}

\vskip 0.3in
]



\printAffiliationsAndNotice{}  

\begin{abstract}
We propose a novel denoising diffusion generative model for predicting nonlinear fluid fields named FluidDiff. By performing a diffusion process, the model is able to learn a complex representation of the high-dimensional dynamic system, and then Langevin sampling is used to generate predictions for the flow state under specified initial conditions. The model is trained with finite, discrete fluid simulation data. We demonstrate that our model has the capacity to model the distribution of simulated training data and that it gives accurate predictions on the test data. Without encoded prior knowledge of the underlying physical system, it shares competitive performance with other deep learning models for fluid prediction, which is promising for investigation on new computational fluid dynamics methods.
\end{abstract}

\section{Introduction}
\label{intro}

Computational fluid dynamics (CFD) is the field that involves the use of numerical techniques to solve the governing equations for fluids, which exhibit chaotic, time-dependent behaviors known as turbulence due to the nonlinear nature of these equations. Despite the nonlinearity, the fine spatial and temporal discretization under high-fidelity conditions is computationally expensive and can consume a large amount of computational resources. All these inconveniences lead to the significant demand for improvements on CFD methodologies.

Data-driven algorithms, represented by deep learning (DL), have received extensive attention in the past decade and have performed well in solving high-dimensional nonlinear problems. Inspired by previous work, we propose an easily trainable generative model for solving nonlinear fluid prediction tasks, which is able to give reasonable predictions without requiring knowledge of the physical laws. At the same time, our model has good flexibility, which makes it only need to be retrained to adapt to different fluid prediction tasks without changing the model structure.

\subsection{Contributions and outline}
The paper contributes by 1) developing the denoising diffusion based fluid flow prediction model FluidDiff; 2) demonstrating the capacity of the model to predict the fluid state at different time points conditional on input data; 3) being competitive with other DL based fluid simulation models for predicting smoke evolution.
The paper is organized as follows. Section \ref{sec:relatedwork} introduces the related work, where the general DL methods and specifically, generative models used in CFD are reviewed, together with diffusion-based generative models; Section \ref{sec:background} describes a general fluid field prediction problem and the principle of denoising diffusion probabilistic model. Section \ref{sec:methodology} shows FluidDiff, where the detailed network architecture and algorithms are presented. Section \ref{sec:experiment} shows the experimental results on a 2D floating-smoke case, together with qualified and quantified discussions; Section \ref{sec:conclusion} concludes the work.

\begin{figure}[ht]
    \centering
    \vskip 0.2in
    \includegraphics[width=\columnwidth]{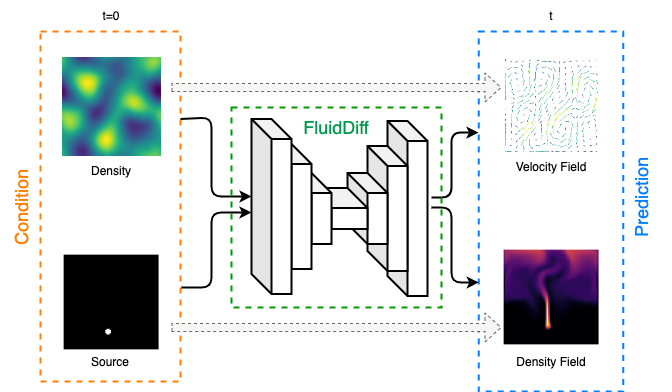}
    \caption{We propose a denoising diffusion generative model for fluid prediction named FluidDiff. The model takes the initial condition of the system and makes predictions of fluid states at a certain time point. The model does not rely on specific physics governing equations, but only on the data gained through numerical simulations, and it can adapt to different tasks without changing the architecture.}
    \label{fig:experiment}
    \vskip 0.2in
\end{figure}

\section{Related Work}
\label{sec:relatedwork}
Various DL-based methods have been applied to enhance the CFD \cite{bar2019learning, kochkov2021machine, li2020fourier, ajuria2020towards, weymouth2022data, ling2016reynolds, jiang2021interpretable, beck2019deep, murata2020nonlinear, rojas2021reduced, eivazi2022towards}. Among them, there are three main areas where the DL can be incorporated with physics-principle-based numerical solvers:
\begin{itemize}
    \item In the direct numerical solver, neural networks (NNs) are widely used to improve the computational efficiency in approximating spatial derivatives \cite{bar2019learning}, finding the correlations between fine and coarse girds \cite{kochkov2021machine}, improving partial differential equations (PDEs) solvers in coarse grid \cite{li2020fourier}, accelerating solving Possion equations \cite{ajuria2020towards} and pressure corrections \cite{weymouth2022data}.
    \item NNs are used to model the turbulence using Reynolds-averaged Navier-Stokes models (RANS) \cite{ling2016reynolds, jiang2021interpretable} and coarsely-resolved large-eddy simulations (LES) \cite{beck2019deep}
    \item DL are also used to develop reduced order models (ROM). More specifically, proper orthogonal decomposition (POD) can be achieved through NNs \cite{murata2020nonlinear, rojas2021reduced, eivazi2022towards}, which is used to map the high-dimensional flow space into a low-dimensional latent space, the latter contains most of the information of system but with much less dimensions.
\end{itemize}

Generative models such as auto-encoders (AEs)\cite{kingma2013auto}, energy-based models (EBMs)\cite{lecun2006tutorial}, normalized flows \cite{dinh2014nice} and generative adversarial networks (GANs)\cite{goodfellow2014generative} can be regarded as a kind of ROM since they are all able to convert original flow spaces to certain latent spaces, the latter usually are of lower dimension. Sampling from the latent spaces leads to predictions. Among them, GANs are the most representative ones due to their outstanding sampling quality and general type. Another merit of GANs is the loss function can be modified to be more physically plausible, which is usually achieved by introducing physical constraints into the loss of generator. Therefore, GANs have attracted the interest of many researchers and been applied to fluid modeling and prediction \cite{farimani2017deep, xie2018tempogan, cheng2020data, akkari2020deep, chu2021learning, yousif2021high, drygala2022generative, wu2022navier, ferreira2022framework, xie2022dualsmoke}. Farimani \textit{et al.}\cite{farimani2017deep} train a conditional GAN (cGAN) to generate the steady solution of heat conduction and impressible fluid flow without any physical knowledge. Xie \textit{et al.}\cite{xie2018tempogan} use a cGAN to address the super-solution problem for spatio-temporal fluid flows, the model can enable highly-detailed velocities or vorticities from low-resolution inputs. Cheng \textit{et al.} develop a deep convolutional GAN (DCGAN) to predict spatio-temporal flow distributions, then apply it to a real-world case and gained good consistency with rapid computational speed. Akkari \textit{et al.} extended Farimani \textit{et al.}\cite{farimani2017deep}'s work, they study the use of DCGAN on impressible unsteady fluid flow in a channel with a moving obstacle inside, the model can not only memorize the training data, but also given new reasonable predictions. Chu \textit{et al.}\cite{chu2021learning} focus on using cGANs to solve the ill-posed problems of fluids, which requires model to derive plausible predictions from sparse inputs, e.g. a single frame of a density field. Also, the predictions are able to be controlled by modalities like obstacles and physical parameters. Yousif \textit{et al.}\cite{yousif2021high} train a GAN that can reconstruct high-resolution turbulent flows with coarse ones. They also test the possibility of using transfer learning in flow prediction tasks, which can be meaningful for reducing the computational cost of CFD. Meanwhile, Drygala \textit{et al.}\cite{drygala2022generative} give the mathematical proof that GANs are able to learn representations of the chaotic systems from limited state snapshots, which provides the mathematical fundamentals of the implementation of GANs. Wu \textit{et al.}\cite{wu2022navier} incorporate Naiver-Stokes (NS) equations into GAN's loss function, which endows model with more physical meanings, and it outperforms the similar ones. Ferreira \textit{et al.}\cite{ferreira2022framework} apply cGANs on simulations of fluid flow in fractured porous media. They propose using cGAN for upscaling the permeability of single fractures, which leads to a substantial reduction of computational time. Xie \textit{et al.}\cite{xie2022dualsmoke} develop a two-stage model based on cGANs that can generate realistic smoke visualizations from hand-written sketches, together with a user-friendly interface that can serve for various design scenarios. Apart from GANs, there are examples of implementations of other generative models on flow prediction tasks. Kim \textit{et al.}\cite{kim2019deep} present a generative model based on AE to synthesize fluid simulations from a set of reduced parameters, which is used for 2D and 3D smoke data. Morton \textit{et al.}\cite{morton2021parameter} also use a varientional AE (VAE) for parameterized fluid prediction tasks.

Diffusion models \cite{hyvarinen2005estimation, vincent2011connection, sohl2015deep} have emerged as a new catalog of deep generative models, which show comparable or even better performance than GANs in various computer vision (CV) tasks \cite{song2019generative, ho2020denoising, dhariwal2021diffusion, ho2022cascaded, rombach2022high, saharia2022palette, gu2022vector, daniels2021score}. In general, diffusion models can be divided into three main categories: (\textit{i}) Denoising diffusion probabilistic models (DDPMs)\cite{ho2020denoising}; (\textit{ii}) Score-based diffusion models\cite{song2019generative}; (\textit{iii}) Stochastic differential equation (SDE) based models \cite{song2020score}, where (\textit{iii}) can be treated as the generalization of (\textit{i}) and (\textit{ii}). Among them, DDPMs and their variants are the most widely used and well studied. However, to the best of our knowledge, despite of the popularity of DDPMs in CV, there is currently no precedent for applying it to fluid flows prediction tasks, the only application is \cite{shu2022physics}, who used DDPM for super-resolution reconstructions of fluid simulation. Compared with its competitor GAN, the biggest advantage of DDPM is that it can avoid training instability and model collapse. The former comes from the game competition between the generator network and the discriminator network in GAN, while the latter is because the generator in GAN only learns a subset of the entire data space. All these two drawbacks of GANs requires researchers to carefully design the network structure and loss function. Nevertheless, the trade-off of stability is the slower sampling speed of DDPM, since the sampling of DDPM is realized by iterative Langevin dynamics, which requires hundreds of inference steps. So far, some research have been done to accelerate the sampling speed\cite{song2020denoising, jolicoeur2021gotta}.

\section{Background}
\label{sec:background}

\subsection{A general fluid flow prediction problem}
Predicting complex fluid flows is usually addressed through solving a nonlinear PDE, such as NS equation, Euler equation, or other equations suit for different circumstances and approximations under some constraints, also with some certain initial and boundary conditions. Generally speaking, they can be ascribed to the following equation set:
\begin{gather}
    \frac{\partial u}{\partial \tau} = f(u, \mathbf{x}, \nabla_x\mathbf{x}, \nabla^2_x\mathbf{x}) \label{eq:fluidEq}\\
    g(u, \mathbf{x}, \tau) = 0 \label{eq:fluidConstraint1}\\
    \varphi_1(u, \mathbf{x}, \tau)|_{\mathbf{x} = \mathbf{x}_0} = 0 \label{eq:fluidConstraint2}\\
    \varphi_2(u, \mathbf{x}, \tau)|_{\tau=0} = 0 \label{eq:fluidConstraint3}
\end{gather}
where $u$ is the physical variable we are interested in, which can be either scalar (e.g. density, pressure) or vector (e.g. velocity). $\mathbf{x}$ is the spatial coordinates, $\nabla_{\mathbf{x}}$ and $\nabla^2_{\mathbf{x}}$ represents the first and second order spatial derivative respectively. $\tau$ is the time, $g$ is the constraints, which can be physical (e.g. impressibility, heat conduction) or artificial (e.g. geometric obstacles, sources). $\varphi_1$, $\varphi_2$ stands for boundary condition and initial condition respectively.

By solving equation \eqref{eq:fluidEq} under constraints \eqref{eq:fluidConstraint1}\eqref{eq:fluidConstraint2}\eqref{eq:fluidConstraint3}, usually numerically, one can receive the solver as a mapping: $u = \hat{u}(\mathbf{x}, \tau)$, Further, if $g$, $\varphi_1$ and $\varphi_2$ can be parameterized to a parameter vector $\mathbf{c}$, then $u = \hat{u}(\mathbf{x}, \tau, \mathbf{c})$. In the conventional way, a variety of numerical methods have been developed to obtain $\hat{u}$. Under some assumptions, the numerical solver can be treated as a mapping $\mathcal{M}: \mathbf{x}, \tau, \mathbf{c} \to \hat{u}(\mathbf{x}, \tau, \mathbf{c})$. A novel approach is to use a data-driven model like neural network to substitute traditional numerical solver, but also reaches $\mathcal{M}$. We demonstrate that, after using dedicated-designed architecture and sufficient training, a DDPM can approximate $\mathcal{M}$ well, so that with a set of given parameters $(\mathbf{x}, \tau, \mathbf{c})$, the model can predict $u$ without solving Equation \eqref{eq:fluidEq}, which demonstrates the principle of fluid flow prediction problem.

\subsection{Denoising diffusion probabilistic model}
DDPM consists of two processes: forward (diffusion) process and reverse (denoising) process. In the forward process, the original training data are perturbed by a series of Gaussian noises with different means and variances in a discrete period $t\in\{1, 2,\dots, T\}$. In each step the corrupted data $x_1, x_2, \dots, x_T$ is Markovian:
\begin{align}
    p(x_t|x_{t-1}) =& \mathcal{N}\left(x_t;\sqrt{1-\beta_t}\cdot x_{t-1}, \beta_t\cdot\mathbf{I}\right) \notag\\
    p(x_t|x_0) =& \mathcal{N}\left(x_t;\sqrt{\hat{\alpha_t}}\cdot x_0, (1-\hat{\alpha}_t)\cdot\mathbf{I}\right)
    \label{eq:ddpmForwardProcess}
\end{align}
where $T$ is the number of diffusion steps, $\beta_1, \dots, \beta_T \in [0,1)$ are hyperparameters that control the variance of noise. $\mathbf{I}$ is the identity matrix with the same dimension as the data. $\mathcal{N}(x;\mu, \sigma)$ stands for normal distribution with mean $\mu$ and standard deviation $\sigma$. $\alpha_t = 1-\beta_t$, $\hat{\alpha}_t = \prod^{t}_{i=1} \alpha_i$ and $\mathcal{U}$ is the uniform distribution. The variance schedule $(\beta_t)^T_{t=1}$ is chosen that $\hat{\beta}_T\to 0$ and therefore $p(x_T) \simeq \mathcal{N}(0, \mathbf{I})$. Moreover, if the diffusion is small enough, i.e. $(\beta_t)^T_{t=1} \ll 1$, the reverse transition probability $p(x_{t-1}|x_t)$ should have the same function form as the forward process, which is also Gaussian
\begin{align}
    q(x_{t-1}|x_t) = \mathcal{N}\left(x_{t-1};\mu_t(x_t, t), \sigma_t(x_t, t)\right)
    \label{eq:ddpmReverseProcess}
\end{align}
which means if one starts from a Gaussian noise $\mathcal{N}(0, \mathbf{I})$ and applies Equation \eqref{eq:ddpmReverseProcess} progressively, it will finally reach the original distribution $p(x_0)$. In \cite{ho2020denoising}, $\sigma_t$ in Equation \eqref{eq:ddpmReverseProcess} is fixed to be a constant and $\mu$ is the function of the clean data $x_0$ and the corrupted data $x_t$ at time $t$:
\begin{gather}
    \mu_t(x_t, x_0) = \frac{\sqrt{\hat{\alpha}_{t-1}}\beta_t}{1-\hat{\alpha}_t}x_0+\frac{\sqrt{\alpha_t}(1-\hat{\alpha}_{t-1})}{1-\hat{\alpha}_t}x_t \notag\\
    \sigma_t = \frac{1-\hat{\alpha}_{t-1}}{1-\hat{\alpha}_t}\beta_t
    \notag
\end{gather}
Maximizing likelihood is used in the training with optimizing the variational lower-bound:
\begin{align}
    \mathcal{L}_{DDPM} &= \mathbb{E}_p\left[-\log\frac{q_\theta(x_{0:T})}{p(x_{1:T}|x_0)}\right] \notag\\
    &=\mathbb{E}_p[D_{KL}(p(x_T|x_0)||p(x_T)) \notag \\
    &\quad+\sum_{t>1}D_{KL}(p(x_{t-1}|x_t,x_0)||q_{\theta}(x_{t-1}|x_t)) \notag \\
    &\quad-\log q_{\theta}(x_0|x_1)] \notag\\
    &=\mathbb{E}_{p} \left[\frac{1}{2\sigma_t}\Vert \mu_t(x_t,x_0) - \mu_{\theta}(x_t, t) \Vert^2\right] +C
    \label{eq:ddpmLoss}
\end{align}
where $D_{KL}$ represents KL divergence. One shall find that maximizing likelihood here is essentially to approximate the mean in the reverse process. With the reparameterizing trick that $x_t(x_0, z) = \sqrt{\hat{\alpha_t}}x_0 + \sqrt{1-\hat{\alpha_t}}z$ for $z\sim\mathcal{N}(0, \mathbf{I})$, Equation \eqref{eq:ddpmLoss} can be adapted to a simpler version:
\begin{gather}
    \mathcal{L}^{simple}_{DDPM} = \mathbb{E}_{x_0}\left[||\epsilon - \epsilon_{\theta}(\sqrt{\hat{\alpha}_t}x_0 + \sqrt{1-\hat{\alpha}_t}\epsilon, t)||^2\right] \notag\\
    x_0\sim p_{data}(x_0),\quad \epsilon\sim\mathcal{N}(0, \mathbf{I})
    \label{eq:ddpmFinalLoss}
\end{gather}
A NN is trained to approximate the noise added on $x_t$. With the noise network, the reverse process can be implemented via sampling process:
\begin{gather}
    x_{t-1} = \frac{1}{\sqrt{\alpha_t}}\left(x_t-\frac{1-\alpha_t}{\sqrt{1-\hat{\alpha}_t}}\epsilon_{\theta}(x_t, t)\right) + \sqrt{\sigma_t} z \notag \\
    \forall t\in \{T, \dots, 1\},\quad z \sim \mathcal{N}(0, \mathbf{I})
    \notag
\end{gather}

\section{Methodology}
\label{sec:methodology}
Suppose the fluid field $x$ satisfies a certain posterior distribution $p(x|\tau, \mathbf{c})$, where $c$ is the parameterized constraints and $\tau$ denotes the actual time for the physical flow. Note that $\tau$ is different from $t$ in the diffusion process. A conditional denoising diffusion model can be trained to approximate $p(x|\tau, \mathbf{c})$ and demanding states can be sampled from it. Unlike the unconditional case \eqref{eq:ddpmFinalLoss}, the posterior is highly dependent on $\tau$ and $c$. For simplicity we ascribe them all to the condition $y$. Now the task is to approximate and sample from $p(x|y)$, which is usually referred as the conditional generation task.

In \cite{rombach2022high}, the author argues that a conditional denoising decoder $\epsilon(x_t, t, y)$ can be used to control the conditional diffusion process through inputs $y$, where $y$ can be texts, semantic maps or other task. They also use cross attention mechanism to deal with multi-modalities. In our case, $y$ are spatial and temporal information of the system that contains two parts, which are modified to fit the $u$, as shown in Figure \ref{fig:model}. The training object of our model is still minimizing the L2 loss between $\epsilon$ and $\epsilon_{\theta}$:
\begin{gather}
    \mathcal{L} = \mathbb{E}_{x_0, y}\left[||\epsilon - \epsilon_{\theta}(\sqrt{\hat{\alpha}_t}x_0 + \sqrt{1-\hat{\alpha}_t}\epsilon, t, y)||^2\right] \notag \\
    x_0, y \sim p_{data}(x_0, y), \quad \epsilon\sim\mathcal{N}(0, \mathbf{I})
    \notag
\end{gather}
where diffusion time step $t$ serves as the indicator of diffusion sequence. We used the same method as the position embedding in\cite{vaswani2017attention}, which is widely used in the Transformer architecture. Through this method, a given scalar $s$ is encoded into a vector $\mathbf{v}$ with the formulation:
\begin{gather}
    v_i(s) =
    \begin{cases}
        \sin (\omega_k s), i=2k \\
        \cos (\omega_k s), i=2k + 1
    \end{cases} \notag \\
    \omega_k = \frac{1}{10000^{2k/d}}
    \label{eq:positionEmbed}
\end{gather}
where $v_i$ is the i-th entry of $\mathbf{v}$ and $d$ is the size of embedding. We use it to encode diffusion step $t$ as a vector $\mathbf{t}$. For the spatial part of $y$, it is depicted as a matrix with the same resolution as the training snapshots $x$. The temporal information $\tau$ is also extended to be a matrix with the same size as $x$. It is full of identical entries that have been normalized with the total simulation time. The noise is added only on $x$ to get $x_t$, finally, we concatenate $x_t$ and $y$ to form a multi-channel input, and feed it together with diffusion time embedding $\mathbf{t}$ into the network.

For the network, we use a U-Net architecture to predict $\epsilon(x_t,t,y)$, please see the right part of Figure \ref{fig:model} for details. Both down and up sampling part consist of 4 main sections, In each section, there are two Resnet-like blocks, which is made up of a $3\times3$ convolutional layer, a group normalization and a SiLU activation, the residual connections also exists although they are not explicitly shown in the figure. Besides the Resnet-like blocks, an embedding MLP is also used to map the fixed-size diffusion step vector into the size that suits for element-wise addition with the output feature maps from Resnet-like blocks. After the addition, the output will pass through a Transformer layer, which will do self-attention to extract important representations, and finally be halved the size by a down-sampling convolutional layer. The bottleneck block is a "sandwich" structure, with two Resnet-like blocks on either side and one Transformer block in the middle, all of them share the same structures as their corresponding ones in the down-sampling path. Along the up-sampling path, the configuration is symmetric to the down sampling one, except the Resnet-like blocks in the up-sampling part not only receive the output from former block, but also the skip connection from the corresponding down-sampling block. The final layer in each section is replaced by a $3\times3$ transposed convolutional layer that doubles the size too. The training and sampling process is implemented by \cref{alg:ddpmTraining} and \cref{alg:ddpmSampling}.

\begin{algorithm}[tb]
   \caption{Training}
   \label{alg:ddpmTraining}
\begin{algorithmic}
\STATE {\bfseries Input:} $x_0\sim p(x_0)$,$y\sim p(y|x_0)$, $T$, $(\hat{\alpha}_t)^T_{t=1}$
\STATE {\bfseries Output:} $\epsilon_{\theta}$
\REPEAT
\STATE $t\sim \mathcal{U}(\{1,\dots,T\})$ \\
\STATE $\epsilon \sim \mathcal{N}(0, \mathbf{I})$ \\
\STATE $\mathcal{L} = ||\epsilon - \epsilon_{\theta}(\sqrt{\hat{\alpha}_t}x_0 + \sqrt{1-\hat{\alpha}_t}\epsilon, t/T, y)||^2$
\STATE Back propagate $\mathcal{L}$ using gradient descent
\UNTIL {converged}
\end{algorithmic}
\end{algorithm}

\begin{algorithm}[tb]
   \caption{Sampling}
   \label{alg:ddpmSampling}
\begin{algorithmic}
\STATE {\bfseries Input:} $\epsilon_{\theta}$, $T$, $(\alpha_t)^T_{t=1}$, $(\hat{\alpha}_t)^T_{t=1}$, $(\sigma_t)^T_{t=1}$
\STATE {\bfseries Output:} $\hat{x}_0\sim \hat{p}(u_0)$
\STATE $u_T \sim \mathcal{N}(0, \mathbf{I})$
\FOR{$t=T,\dots, 1$}
\STATE $z\sim \mathcal{N}(0, \mathbf{I})$ if $t>1$ else $z=0$ \\
\STATE $x_{t-1} = \frac{1}{\sqrt{\alpha_t}}\left(x_t-\frac{1-\alpha_t}{\sqrt{1-\hat{\alpha}_t}}\epsilon_{\theta}(x_t, t/T, y)\right) + \sqrt{\sigma_t} z$
\ENDFOR
\end{algorithmic}
\end{algorithm}

 The form of $y$ can be more diverse, as long as it can be represented as grid data, like geometric obstacles, special boundary conditions etc. If the problem is more specific, other types of conditions like forces, temperature, pressure, viscosity can also be introduced as additional channels. This feature enable our model with great flexibility for different fluid simulation prediction tasks.

\section{Experiment}
\label{sec:experiment}
\subsection{Problem setup}
We tested the model on a 2D random floating-smoke case. In this case, a domain with size of $64\times64$ is full of randomly distributed still smoke. The smoke is free to float upward by a constant buoyancy force in the vertical direction, the process is shown in Figure \ref{fig:experiment}. The boundaries are closed in all directions. The movement of smoke can be described by it's velocity field $\vec{u}$, which is governed by the incompressible Navier-Stokes equations:
\begin{gather}
    \nabla\cdot \vec{u} = 0 \notag\\
    \frac{\partial \vec{u}}{\partial t} + (\vec{u}\cdot\nabla)\vec{u} = - \nabla p + \nu\nabla^2 \vec{u} + \eta \vec{d} \notag \\
    \vec{u}(x, y, 0) = 0 \notag\\
    \vec{u}(x_{boundary}, y_{boundary}, t) = 0 \notag
\end{gather}
where $p$ is the pressure, $\vec{d}$ is the unit direction vector towards y direction, $\nu$ is the viscosity coefficient and $\eta$ is a constant to approximate the Boussinesq buoyancy. The task is to predict $\vec{u}$ at a certain time with a given initial density configuration which will decide the unique $p$. These equations are solved using semi-Lagrangian method under $\phi_{\text{Flow}}$ \cite{phiflow} framework to gather training data. Specifically, we start with 1000 random density scenes. The property of smoke is set as $\nu=0.03$ and $\eta=0.5$. The total simulation time is chosen as 40.0s with sampling time step of 1.0s, i.e. 40 snapshots for each experiment. The x and y components of velocity are recorded as $64\times64$ grid data. Therefore, there are 40000 snapshots for both $u_x$ and $u_y$. During the training phase, we randomly split 1000 experiments into training and test groups in a 4:1 ratio, each group contains complete 40 snapshots for the whole simulation period. i.e. 32000 samples for the training set and 8000 samples for the test set. The total amount of diffusion steps is 400 and $\beta$ is chosen within $[0.0001, 0.02]$ as suggested in \cite{ho2020denoising}. The optimizer is Adam with learning rate of 0.0001, a cosines decay is also used to adjust the learning rate. The training is performed for 40 epochs with the batch size of 8 on a GTX Titan Xp GPU for 10 hours.
\begin{figure}[ht]
    \centering
    \vskip 0.2in
    \includegraphics[width=\columnwidth]{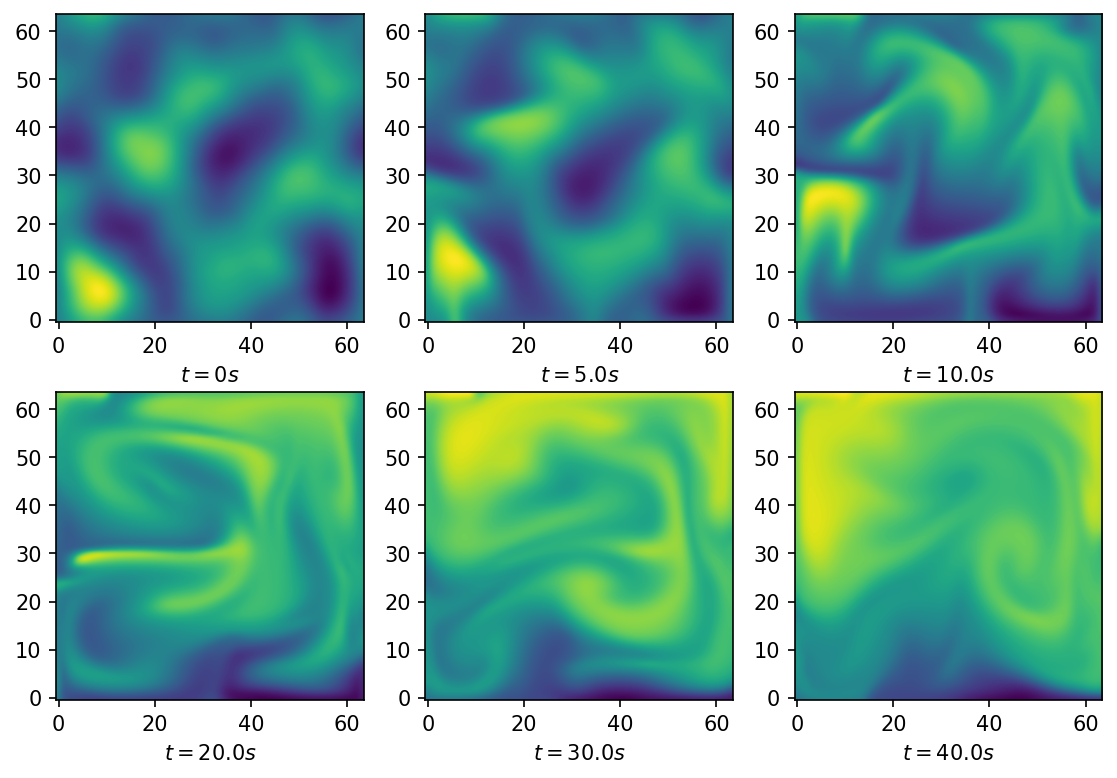}
    \caption{The floating-smoke experiment, the incompressible smoke is floating freely within a closed domain under a constant buoyancy force. The initial density is distributed randomly in each sub experiment. One of the sub experiments is shown here, the evolution of density is depicted while the velocity fields are omitted.}
    \label{fig:experiment}
    \vskip 0.2in
\end{figure}

\subsection{Reproducing the training data}
Since our model does not directly output the velocity field, it is worth testing the modelling capacity of the model on the training set first. Figure \ref{fig:train_pred} shows the predictions of one experiment in the training set. The absolute errors are also exhibited in Figure \ref{fig:train_error}. In order to verify the model's capacity of learning the distribution of data, we further investigate the probability density distribution of the predictions, the results can be seen in Figure \ref{fig:train_hist}. These results shows that the model indeed captures the distribution of training data. However, with the increase of time, the velocity field becomes more and more irregular, and the error of the prediction result also increases gradually. We suspect that this is due to the fact that in the physics problem we study, different experiments with different initial conditions tend to evolve to similar states at the end of simulation, and the model can easily confuse them and ignore the nuances. Due to the lack of regularization based on physical laws and obvious prior guidance, our model produces large long-term prediction errors. Different from traditional CV tasks that require sample diversity, we want to generate single and accurate results. We hypothesize that the problem can be mitigated by introducing guidance based on physical laws during the training phase.

\subsection{Predictions on the test set}
Since the initialization of the experiments is random, we assume the samples in the test set are independent of the training set and can be used to test the generalization performance of our model. Figure \ref{fig:test_pred}, \ref{fig:test_error} and \ref{fig:test_hist} show the prediction results, absolute error and probability density distribution respectively. As with the training set, on the test set, the model makes reasonable predictions in the short term, but has larger errors in the long term. This point is also reflected in the probability density distribution. We believe that this points to our model having less ability to capture fine differences between samples. When samples are far apart in the data distribution, i.e. the early stage of fluid evolution, where the evolution states of different initial states have obvious distinguishing features, the prior is able to guide the denoising process towards the correct target predictions efficiently. However, when the samples are very close at the end of evolution, the samples may tend to be homogeneous. Due to the high dimensional nature of the problem, the guidance from prior is weak and the model can easily confuse the predictions which are actually from the wrong initial condition, resulting in poor performance on the test set. At the same time, since the model is not restricted by any physics laws, it occasionally produces unreasonable predictions that violate physics, such as the sudden appearance of large velocity $y$ components at $t=40$ s.

\subsection{Comparison with other models}
\begin{figure}[ht]
    \centering
    \includegraphics[width=\columnwidth]{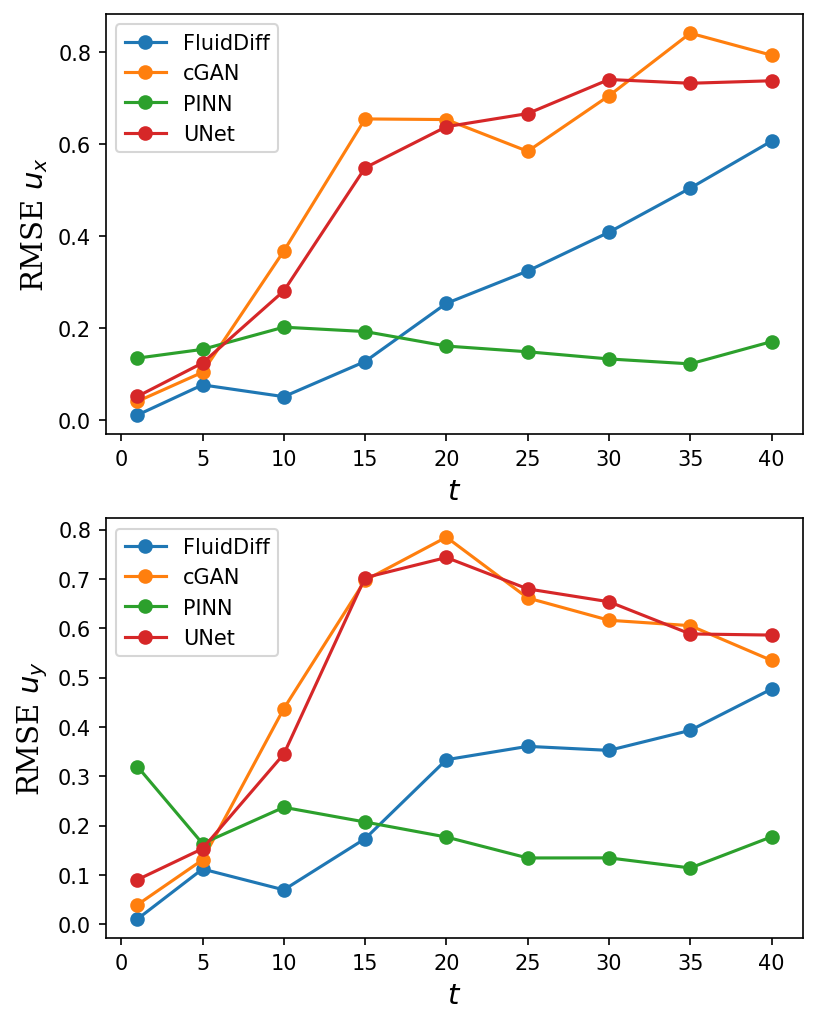}
    \caption{The RMSE of predictions from different models at different time. Except for PINN, the other three models all have the problem of RMSE increasing with the prediction time.}
    \label{fig:test_rmse}
\end{figure}

We also compare the performance of our model with three other models commonly used in CFD prediction tasks. We were not able to find open-source available implementations of these three models specific to the studied problem. Therefore, we have implemented the models to the best of our ability closely following the relevant literature. The three models for the benchmark are: 1) cGAN, consisting of a U-Net as generator and a PatchGAN discriminator 2) PINN, a ten-layer fully-connected network that learns the solution $\vec{u}(x, y, t)$ 3) U-Net, for directly learning the mapping from $y$ to $u$, and $L2$ loss is used in the optimization process. It should be pointed out that in the training process of PINN, since it is different from the training data type of other models (the input of PINN is the (x, y, t) coordinate tuple), only the data from a single experiment are used. We flatten 2D grid data into 1D vector with the length of $64\times64\times40$ and randomly select 10\% of them as the training set, and the remaining 90\% as the test set. Thus, comparison between PINN and the other models is only indicative. Figure \ref{fig:test_u_models} and \ref{fig:test_v_models} show the qualitative comparisons among models and \cref{tb:error} gives the MAE and RMSE of the predictions. In addition, Figure \ref{fig:test_rmse} shows more detailed RMSE at different time points. The PINN can produce the most accurate predictions because it not only relies on the governing equations, but also excludes the initial condition. It benefits from the encoded knowledge of the underlying physics, and it is thus not applicable when the governing equations are unknown. Also when the different complex initial conditions are considered, a PINN can be hard to train. cGANs suffer from the instability of training and mode collapse, which indeed brought challenges to our benchmarks: It was hard to avoid mode collapse. For the U-Net, it shows limitations on predicting samples outside the training set. Our model has the merit to generate relatively accurate predictions without knowing the governing equations, being easy to train (no mode collapse), and it shows good generalization ability in velocity predictions.
\begin{table}[t]
\caption{Prediction errors from different models, a smaller number means a more accurate prediction}
\label{tb:error}
\vskip 0.15in
\begin{center}
\begin{small}
\begin{sc}
\begin{tabular}{lcccr}
\toprule
Model & MAE & RMSE \\
\midrule
cGAN    & 0.4030& 0.5749  \\
PINN    & 0.1324& 0.1767 \\
U-Net   & 0.3894& 0.5603 \\
FluidDiff    & 0.1975 & 0.3137  \\
\bottomrule
\end{tabular}
\end{sc}
\end{small}
\end{center}
\vskip -0.1in
\end{table}

\section{Conclusion}

\label{sec:conclusion}
In this paper, we have proposed a denoising diffusion generative model for fluid prediction, and applied it to a 2D smoke-floating scenario. We shows it is possible to implement diffusion-based generative models on computational fluid dynamics field. The most significant advantage over popular-used GANs is that they are easy to train while maintaining high quality generation performance, which make it a potential competitor.
However, the limitations still exists. The main drawbacks of denoising diffusion generative model lies on three: 1) low sampling speed, the sampling speed highly dependents on the number of diffusion steps, while a sufficiently large number of steps is usually a guarantee of high-quality sampling results. Recently, many improvements have been proposed to improve the sampling speed of diffusion models, which can be used to solve this problem. 2) spatial inaccuracy, since diffusion-based model is originally developed for image-like spatial invariant data, which is contradicted with the accurate fluid predictions. Additional improvements on spatial information should be considered to generate more controllable results. 3) absence of physics constraints, The lack of guidance with practical physical meaning in the generation process can lead the model to give predictions that seriously violate the laws of physics, although this phenomenon is quite rare in our experiments so far. We hypothesize that adding physical guidance to the generation process can address this issue. We look forward to seeing more applications of diffusion-based generative models to enhance CFD.

\bibliography{paper}
\bibliographystyle{icml2023}

\newpage
\appendix
\onecolumn
\section{Model Architecture}
\begin{figure}[htbp]
\begin{center}
\centerline{\includegraphics[width=0.95\textwidth]{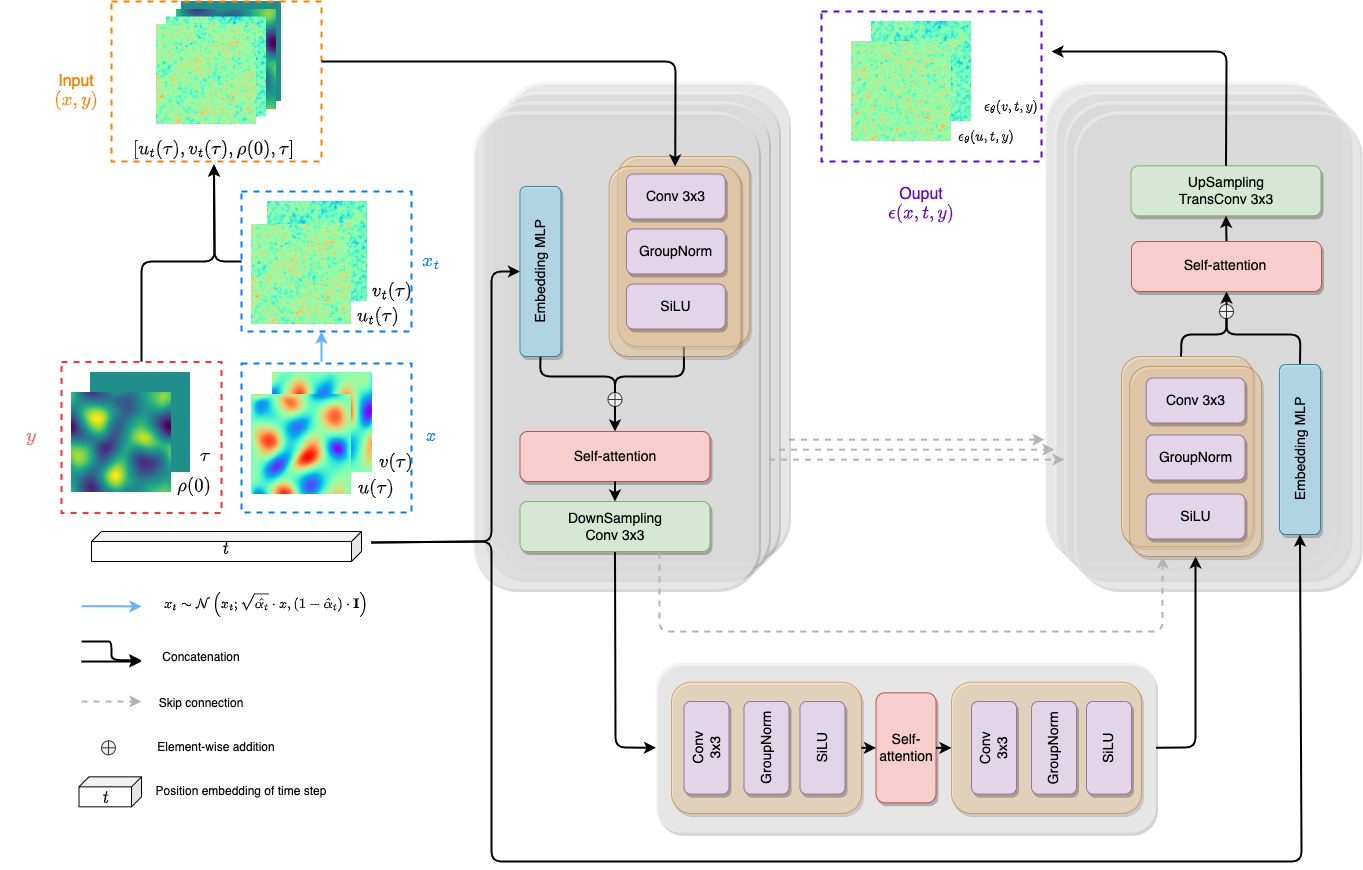}}
\caption{An overview of our proposed model, where the fluid field data samples $x$ are corrupted via diffusion process \eqref{eq:ddpmForwardProcess} to be $x_t$, while initial condition and desired predicting time are ascribed as $y$. Then $x_t$ and $y$ are concatenated as the input of the network, together with the diffusion time step embedding $t$ through \eqref{eq:positionEmbed}. The network is trained to predict the noise $\epsilon$ added on $x$ given $t, y$ and further used in sampling process, which is not shown here}
\label{fig:model}
\end{center}
\end{figure}

\newpage
\section{Experiment}
\begin{figure}[htbp]
    \centering
    \includegraphics[width=0.95\textwidth]{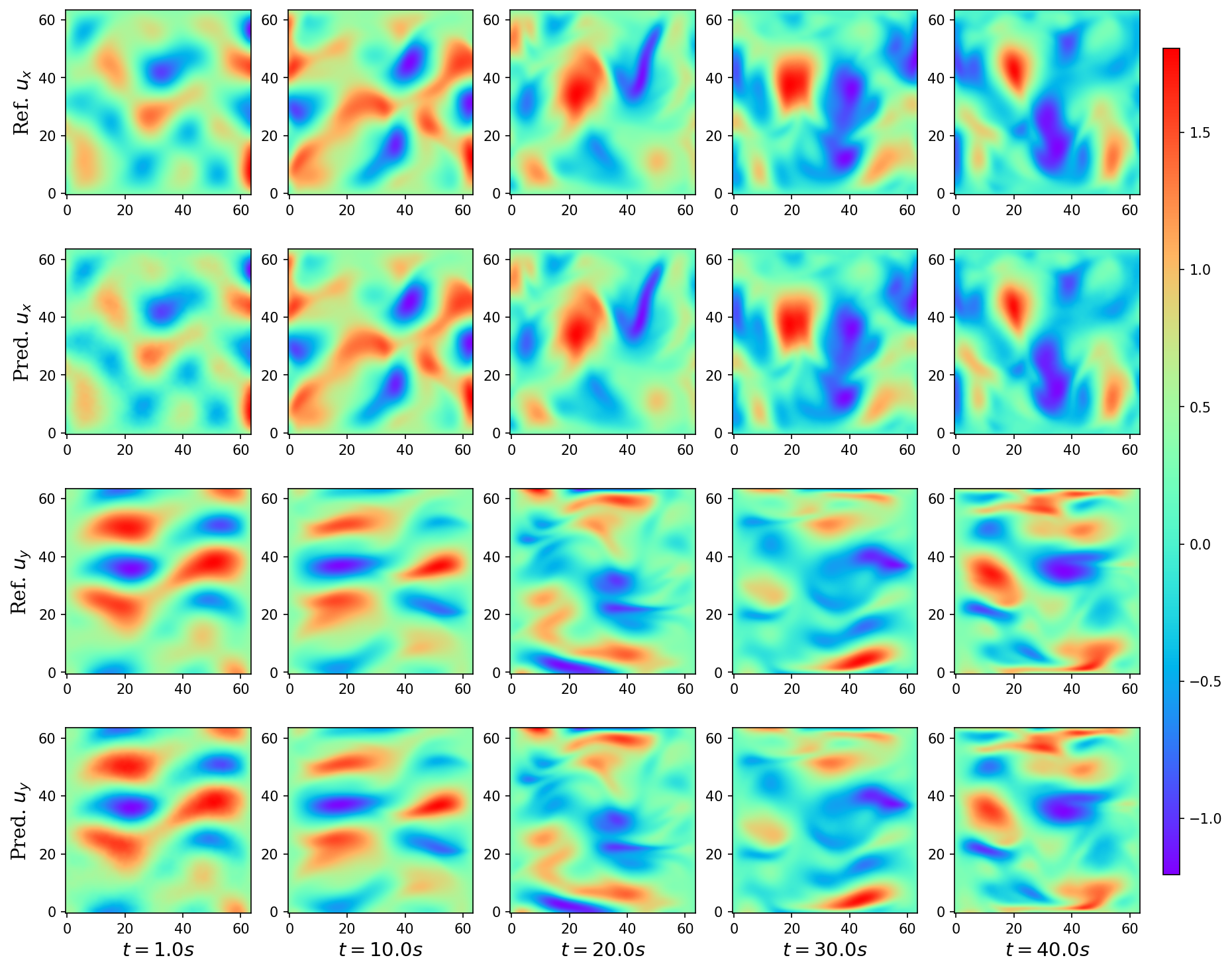}
    \caption{The velocity field predictions of a random experiment in the training set}
    \label{fig:train_pred}
\end{figure}
\begin{figure}[htbp]
    \centering
    \includegraphics[width=0.95\textwidth]{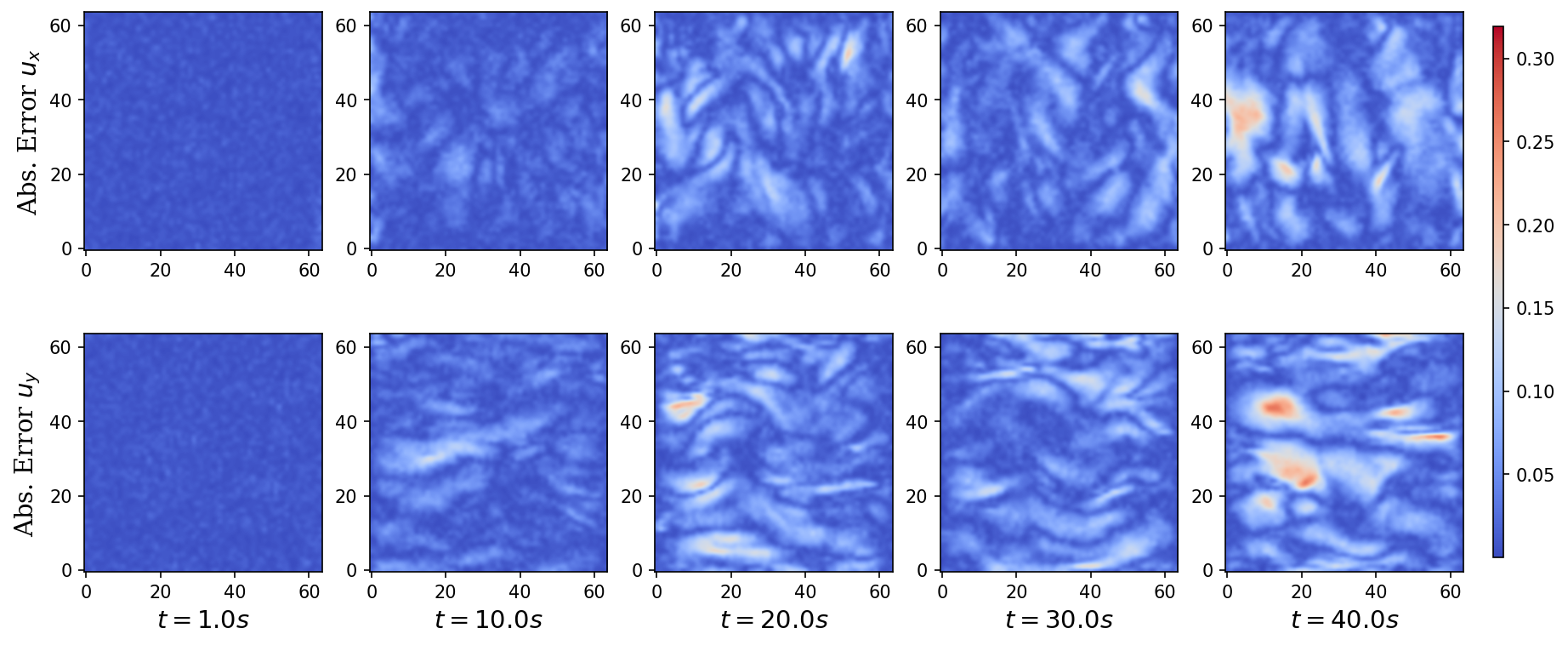}
    \caption{The absolute velocity field prediction errors of a random experiment in the training set.}
    \label{fig:train_error}
\end{figure}
\begin{figure}[htbp]
    \centering
    \includegraphics[width=0.95\textwidth]{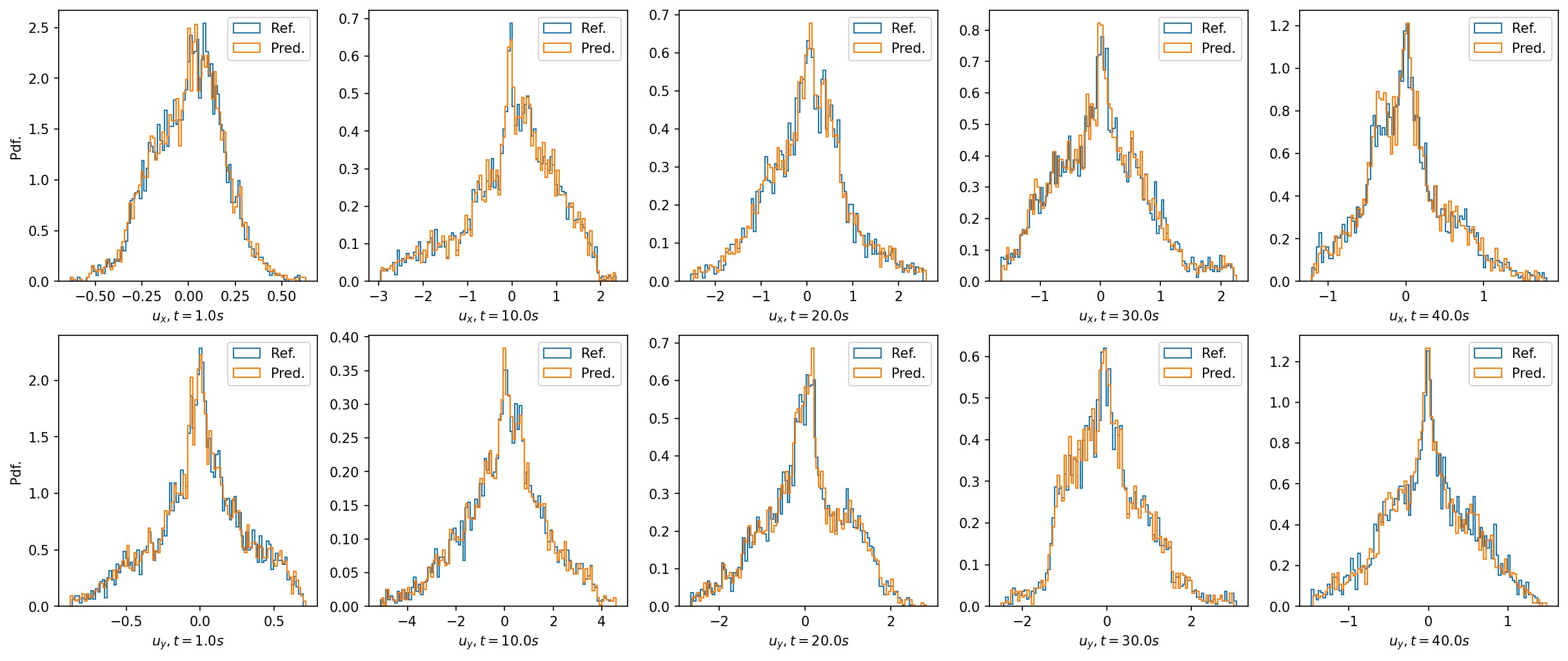}
    \caption{The probability density distribution of velocity field predictions of a random experiment in the training set. Our model succeeds in learning the data distribution of the training set.}
    \label{fig:train_hist}
\end{figure}
\begin{figure}[htbp]
    \centering
    \includegraphics[width=0.95\textwidth]{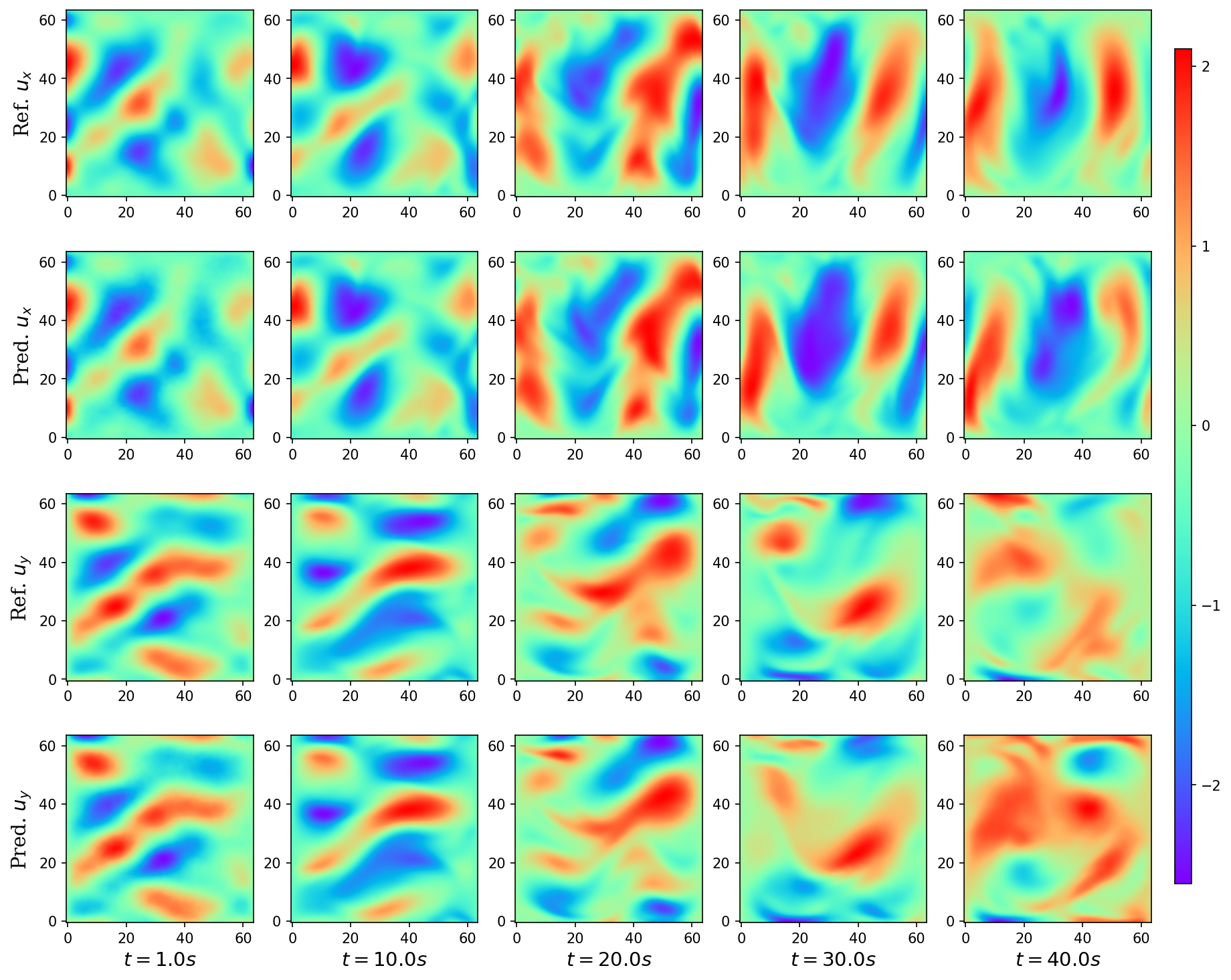}
    \caption{The velocity field predictions of a random experiment in the training set}
    \label{fig:test_pred}
\end{figure}
\begin{figure}[htbp]
    \centering
    \includegraphics[width=0.95\textwidth]{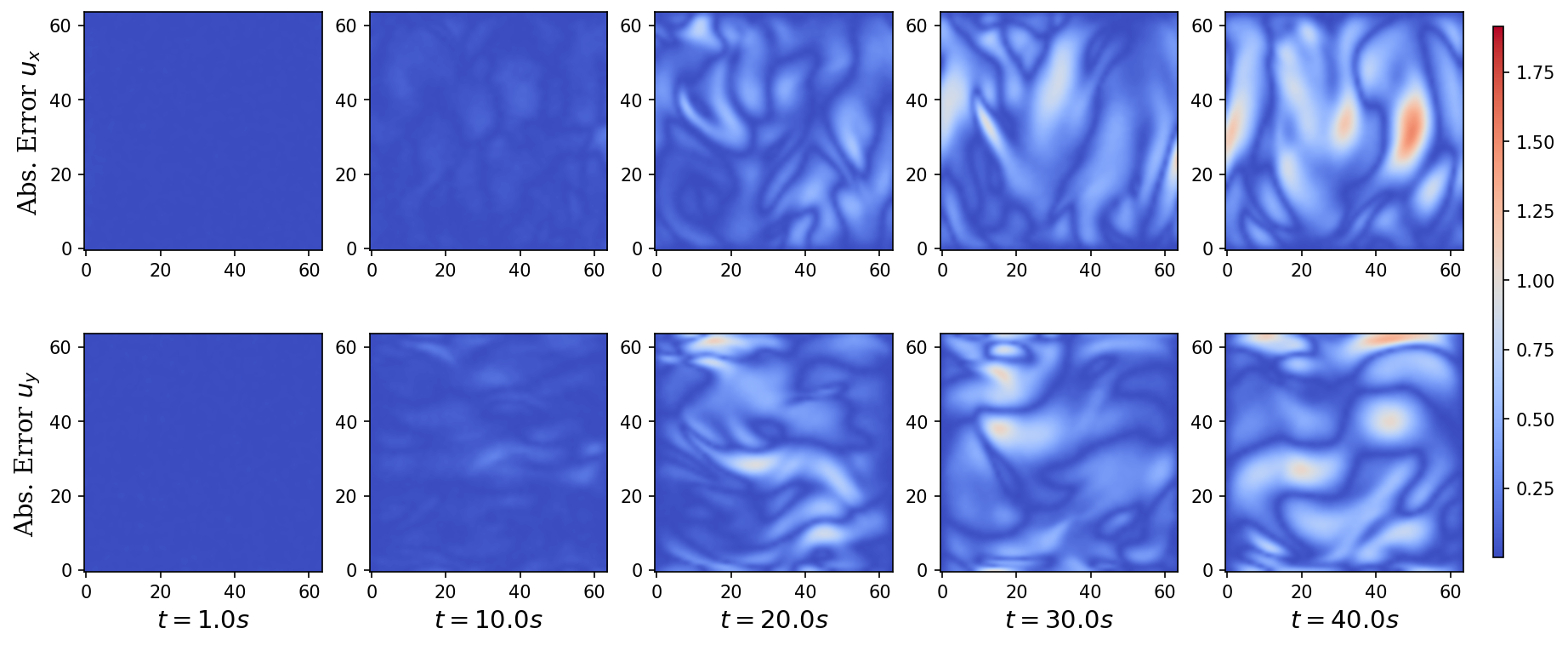}
    \caption{The absolute velocity field prediction errors of a random experiment in the test set}
    \label{fig:test_error}
\end{figure}
\begin{figure}[htbp]
    \centering
    \includegraphics[width=0.95\textwidth]{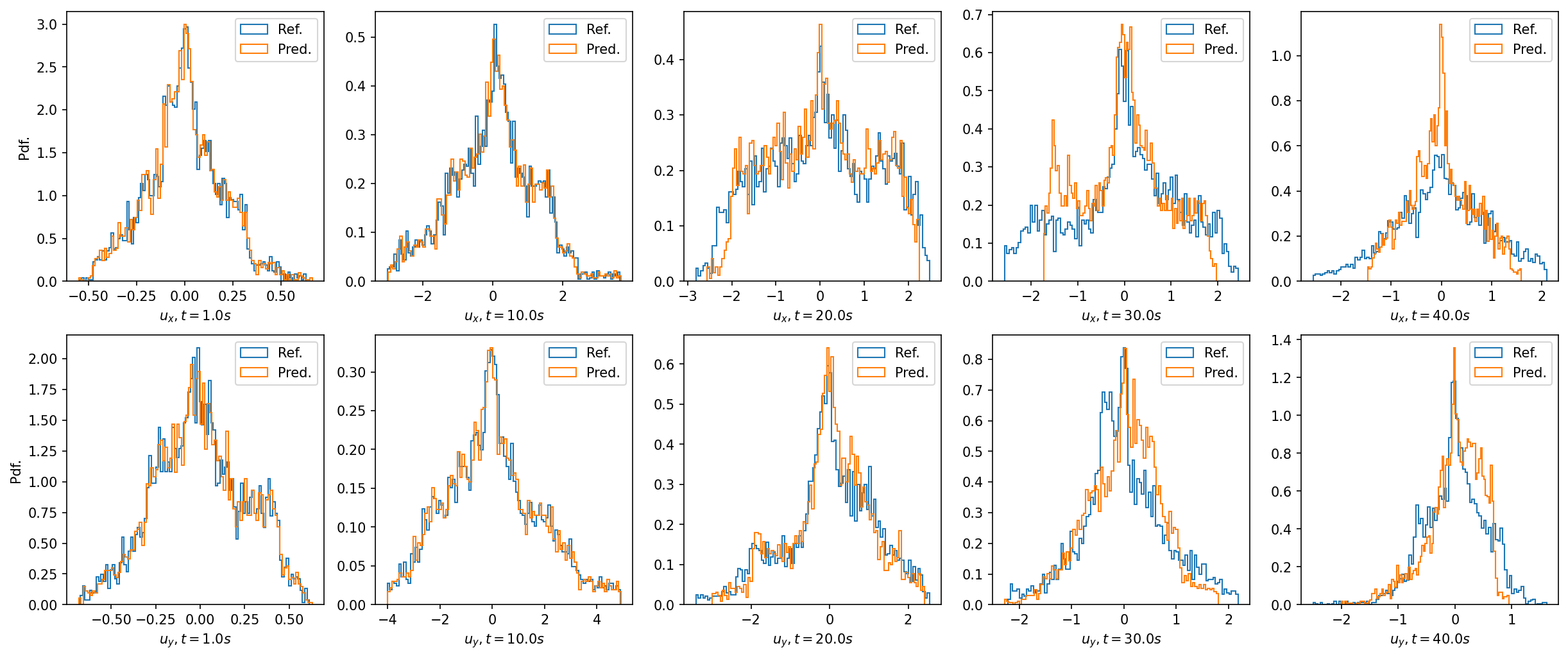}
    \caption{The probability density distribution of velocity field predictions of a random experiment in the test set. In the short term, our model can still capture a relatively accurate velocity distribution, but there is a deviation in the long-term prediction.}
    \label{fig:test_hist}
\end{figure}
\begin{figure}[htbp]
    \centering
    \includegraphics[width=0.95\textwidth]{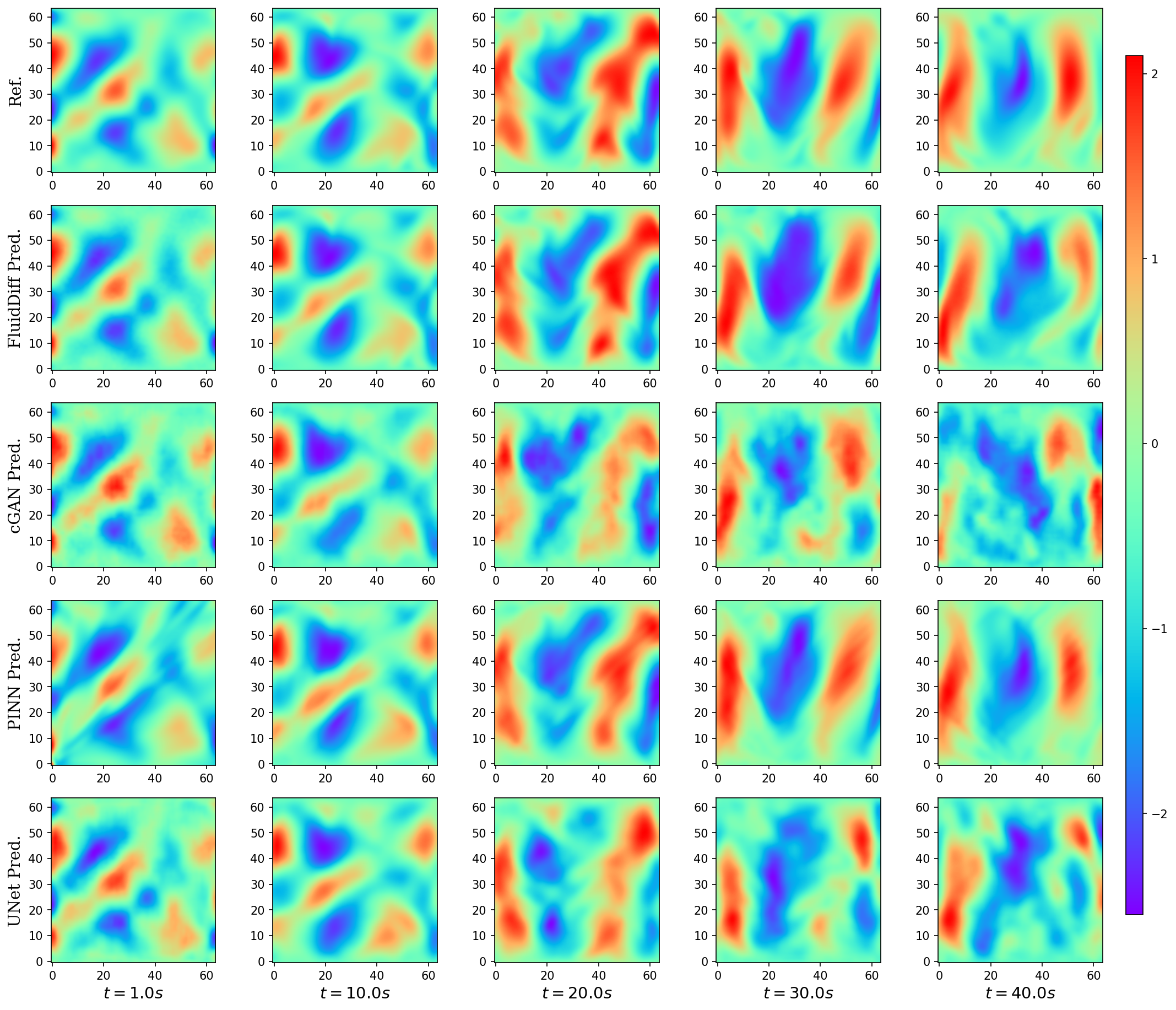}
    \caption{Qualitative comparison of velocity x component predictions from different models. Compared with other models, ours gives reasonable results on x component, and its performance lies second only to PINN. Considering that PINN does not generalize to initial conditions in our benchmark, it can be said that in the test task, our model performs quite well. }
    \label{fig:test_u_models}
\end{figure}
\begin{figure}[htbp]
    \centering
    \includegraphics[width=0.95\textwidth]{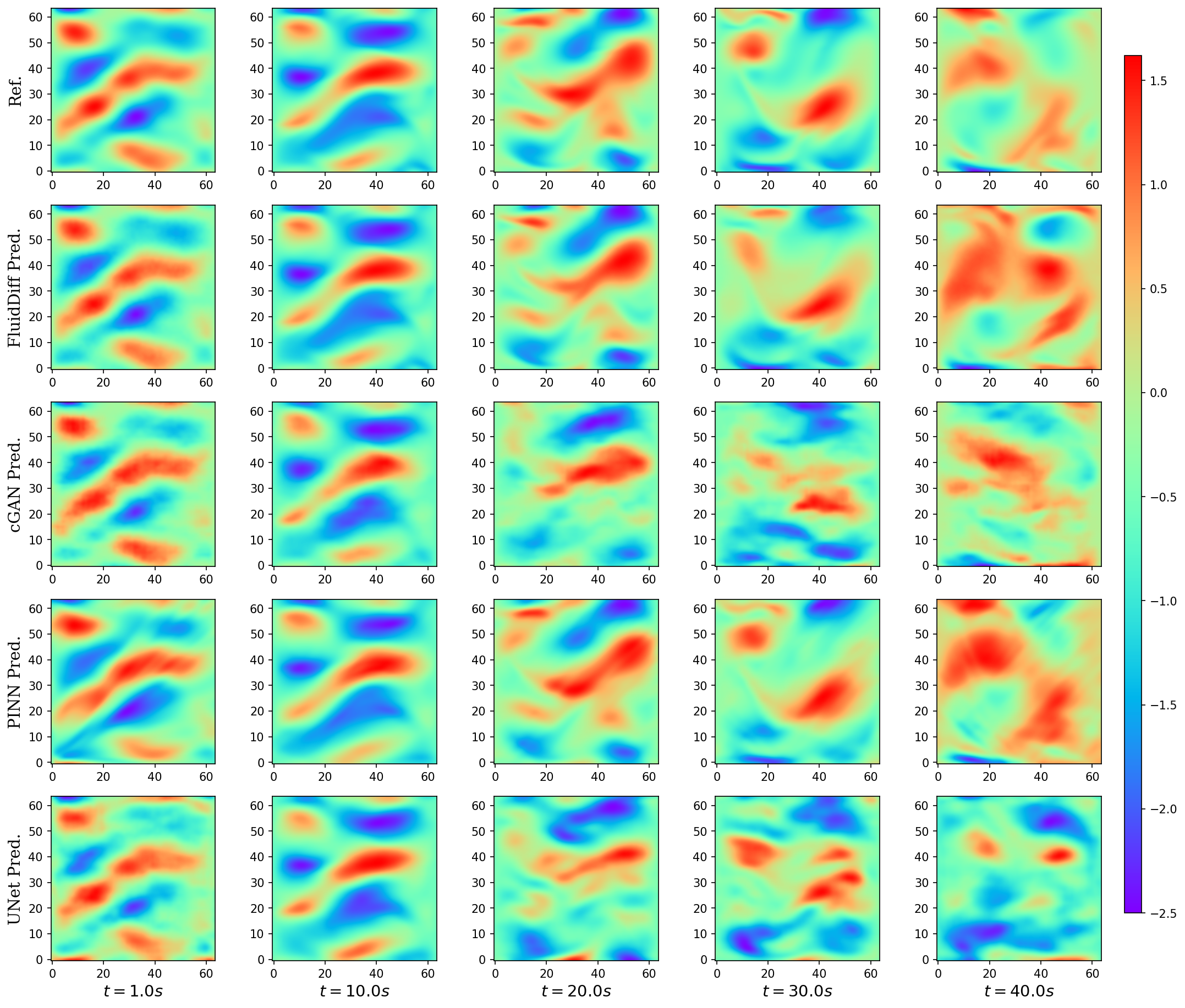}
    \caption{Qualitative comparison of velocity y component predictions from different models. Although FluidDiff gives unreasonable results when predicting the later y component, the overall performance of our model is still good compared with cGAN and U-Net }
    \label{fig:test_v_models}
\end{figure}

\end{document}